# Title: All-Optical Machine Learning Using Diffractive Deep Neural Networks

**Authors: Xing Lin**[1,2,3,†]**, Yair Rivenson**[1,2,3,†]**, Nezih T. Yardimci**[1,3]**, Muhammed Veli**[1,2,3]**, Mona Jarrahi**[1,3] **and Aydogan Ozcan**[1,2,3,4,*]

**Affiliations:**

[1]Electrical and Computer Engineering Department, University of California, Los Angeles, CA, 90095, USA

[2]Bioengineering Department, University of California, Los Angeles, CA, 90095, USA

[3]California NanoSystems Institute (CNSI), University of California, Los Angeles, CA, 90095, USA

[4]Department of Surgery, David Geffen School of Medicine, University of California, Los Angeles, CA, 90095, USA

[*]Correspondence to: ozcan@ucla.edu

[†]These authors contributed equally to this work.

**Abstract:** We introduce an all-optical Diffractive Deep Neural Network (D[2]NN) architecture that can learn to implement various functions after deep learning-based design of passive diffractive layers that work collectively. We experimentally demonstrated the success of this framework by creating 3D-printed D[2]NNs that learned to implement handwritten digit classification and the function of an imaging lens at terahertz spectrum. With the existing plethora of 3D-printing and other lithographic fabrication methods as well as spatial-light-modulators, this all-optical deep learning framework can perform, at the speed of light, various complex functions that computer-based neural networks can implement, and will find applications in all-optical image analysis, feature detection and object classification, also enabling new camera designs and optical components that can learn to perform unique tasks using D[2]NNs.

# Main Text

Deep learning is one of the fastest-growing machine learning methods of this decade (*1*), and it uses multi-layered artificial neural networks implemented in a computer to digitally learn data representation and abstraction, and perform advanced tasks, comparable to or even superior than the performance of human experts. Recent examples where deep learning has made major advances in machine learning include medical image analysis (*2*), speech recognition (*3*), language translation (*4*), image classification (*5*), among many others (*1, 6*). Beyond some of these mainstream applications, deep learning methods are also being used for solving inverse imaging problems (*7-13*).

Here we introduce an all-optical deep learning framework, where the neural network is physically formed by multiple layers of diffractive surfaces that work in collaboration to optically perform an arbitrary function that the network can statistically learn. We term this framework as Diffractive Deep Neural Network ($D^2NN$) and demonstrate its learning capabilities through both simulations and experiments. A $D^2NN$ can be physically created by using several transmissive and/or reflective layers, where each point on a given layer represents an artificial neuron that is connected to other neurons of the following layers through optical diffraction (see Fig. 1A). Following the Huygens' Principle, our terminology is based on the fact that each point on a given layer acts as a secondary source of a wave, the amplitude and phase of which are determined by the product of the input wave and the complex-valued transmission or reflection coefficient at that point. Therefore, an artificial neuron in a diffractive deep neural network is connected to other neurons of the following layer through a secondary wave that is modulated in amplitude and phase by both the input interference pattern created by the earlier layers and the local transmission/reflection coefficient at that point. As an analogy to standard deep neural networks (see Fig. 1D), one can consider the transmission/reflection coefficient of each point/neuron as a multiplicative "bias" term, which is a learnable network parameter that is iteratively adjusted during the training process of the diffractive network, using an error back-propagation method. After this numerical training phase implemented in a computer, the $D^2NN$ design is fixed and the transmission/reflection coefficients of the neurons of all the layers are determined. This $D^2NN$ design, once physically fabricated using e.g., 3D-printing, lithography, etc., can then perform, at the speed of light propagation, the specific task that it is trained for, using only optical diffraction and passive optical components/layers, creating an efficient and fast way of implementing machine learning tasks.

To experimentally validate this all-optical deep learning framework, we used terahertz (THz) part of the electromagnetic spectrum and a standard 3D-printer to fabricate different layers that collectively learn to perform unique functions using optical diffraction through D$^2$NNs. In our analysis and experiments, we focused on transmissive D$^2$NN architectures (Fig. 1) with phase-only modulation at each layer. For example, using five 3D-printed transmission layers, containing a total of 0.2 million neurons and ~8.0 billion connections that are trained using deep learning, we experimentally demonstrated the function of a handwritten digit classifier. This D$^2$NN design formed a fully-connected network and achieved 91.75% classification accuracy on MNIST dataset (*14*) and was also validated experimentally. Furthermore, based on the same D$^2$NN framework, but using different 3D-printed layers, we experimentally demonstrated the function of an imaging lens using five transmissive layers with 0.45 million neurons that are stacked in 3D. This second 3D-printed D$^2$NN was much more compact in depth, with 4 mm axial spacing between successive network layers, and that is why it had much smaller number of connections (<0.1 billion) among neurons compared to the digit classifier D$^2$NN, which had 30 mm axial spacing between its layers. While there are various other powerful methods to achieve handwritten digit classification and design lenses, the main point of these results is the introduction of the diffractive neural network as an all-optical machine learning engine that is scalable and power-efficient to implement various functions using *passive* optical components, which present large degrees of freedom that can be learned through training data.

Optical implementation of learning in artificial neural networks is promising due to the parallel computing capability and power efficiency of optical systems (*15-17*). Being quite different in its physical operation principles compared to previous opto-electronics based deep neural networks (*15, 18-20*), the D$^2$NN framework provides a unique all-optical deep learning engine that efficiently operates at the speed of light using *passive* components and optical diffraction. Benefiting from various high-throughput and wide-area 3D fabrication methods used for engineering micro- and nano-scale structures, this approach can scale up the number of neurons and layers of the network to cover e.g., tens of millions of neurons and hundreds of billions of connections over large area optical components, and can optically implement various deep learning tasks that involve e.g., image analysis (*2*), feature detection (*21*), or object classification (*22*). Furthermore, it can be used to design optical components with new functionalities, and might lead to new camera or microscope designs that can perform various unique imaging tasks that can be statistically learned through diffractive neural networks. As we discuss later in the manuscript, reconfigurable D$^2$NNs can also be implemented using various transmission or reflection based spatial light modulators, providing a flexible optical neural network design that can learn an arbitrary function, be used for transfer learning, and further improve or adjust its performance as needed by e.g., new data or user feedback.

## Results

**D$^2$NN Architecture.** The architecture of a D$^2$NN is depicted in Fig. 1. Although a D$^2$NN can be implemented in transmission or reflection modes by using multiple layers of diffractive surfaces, without loss of generality here we focus on coherent transmissive networks with phase-only modulation at each layer, which is approximated as a thin optical element. In this case, each layer of the D$^2$NN modulates the wavefront of the transmitted field through the phase values (i.e., biases) of its neurons. Following the Rayleigh-Sommerfeld diffraction equation, one can consider every single neuron of a given D$^2$NN layer as a secondary source of a wave that is composed of the following optical mode (*23*):

$$w_i^l(x, y, z) = \frac{z - z_i}{r^2} \left( \frac{1}{2\pi r} + \frac{1}{j\lambda} \right) exp\left( \frac{j2\pi r}{\lambda} \right), \tag{1}$$

where $l$ represents the $l$-th layer of the network, $i$ represents the $i$-th neuron located at $(x_i, y_i, z_i)$ of layer $l$, $\lambda$ is the illumination wavelength, $r = \sqrt{(x - x_i)^2 + (y - y_i)^2 + (z - z_i)^2}$ and $j = \sqrt{-1}$. The amplitude and relative phase of this secondary wave are determined by the product of the input wave to the neuron and its transmission coefficient ($t$), both of which are complex-valued functions. Based on this, for the $l$-th layer of the network, one can write the *output function* ($n_i^l$) of the $i$-th neuron located at ($x_i, y_i, z_i$) as:

$$n_i^l(x, y, z) = w_i^l(x, y, z) \cdot t_i^l(x_i, y_i, z_i) \cdot \sum_k n_k^{l-1}(x_i, y_i, z_i) = w_i^l(x, y, z) \cdot |A| \cdot e^{j\Delta\theta} \,, \tag{2}$$

where we define $m_i^l(x_i, y_i, z_i) = \sum_k n_k^{l-1}(x_i, y_i, z_i)$ as the input wave to $i$-th neuron of layer $l$, $|A|$ refers to the relative amplitude of the secondary wave, and $\Delta\theta$ refers to the additional phase delay that the secondary wave encounters due to the input wave to the neuron and its transmission coefficient. These secondary waves diffract between the layers and interfere with each other forming a complex wave at the surface of the next layer, feeding its neurons. The transmission coefficient of a neuron is composed of amplitude and phase terms, i.e., $t_i^l(x_i, y_i, z_i) = a_i^l(x_i, y_i, z_i) exp(j\phi_i^l(x_i, y_i, z_i))$, and since we focus on a phase-only D²NN architecture the amplitude $a_i^l(x_i, y_i, z_i)$ is assumed to be a constant, ideally 1, ignoring the optical losses – a topic that we expand in the Discussion section. Through deep learning, the phase values of the neurons of each layer of the diffractive network are iteratively adjusted (trained) to perform a specific function by feeding training data at the input layer and then computing the network's output through optical diffraction. Based on the calculated error with respect to the target output, determined by the desired function, the network structure and its neuron phase values are optimized using an error back-propagation algorithm, which is based on the stochastic gradient descent approach used in conventional deep learning (see the Methods section for further details).

Compared to standard deep neural networks, a D²NN is not only different in that it is a physical and all-optical deep network, but also it possesses some unique architectural differences. First, the inputs for neurons are complex-valued, determined by wave interference and a multiplicative bias, i.e., the transmission/reflection coefficient. Second, the individual function of a neuron is the phase and amplitude modulation of its input to output a secondary wave, unlike e.g., a sigmoid, a rectified linear unit (ReLU) or other nonlinear neuron functions used in modern deep neural networks. Third, each neuron's output is coupled to the neurons of the next layer through wave propagation and coherent (or partially-coherent) interference, providing a unique form of interconnectivity within the network. For example, the way that a D²NN adjusts its receptive field, which is a parameter used in convolutional neural networks, is quite different than the traditional neural networks, and is based on the axial spacing between different network layers, the signal-to-noise ratio (SNR) at the output layer as well as the spatial and temporal coherence properties of the illumination source. The secondary wave of each neuron will in theory diffract in all angles, affecting in principle all the neurons of the following layer. However, for a given spacing between the successive layers, the intensity of the wave from a neuron will decay below the detection noise floor after a certain propagation distance; the radius of this propagation distance at the next layer practically sets the receptive field of a diffractive neural network and can be physically adjusted by changing the spacing between the network layers, the intensity of the input optical beam, the detection SNR or the coherence length and diameter of the illumination source.

**D²NN trained for handwritten digit classification.** We first trained a $D^2NN$ as a digit classifier to perform automated classification of handwritten digits, from zero to nine (see Fig. 1B and Fig. 2A). For this task, phase-only transmission masks were designed by training a 5-layer $D^2NN$ with ~55,000 images from MNIST handwritten digit database (*14*). Input digits were encoded into the amplitude of the input field to the $D^2NN$, and the diffractive network was trained to map input digits into ten detector regions, one for each digit (0,1,…,9). The classification criterion was to find the detector that has the maximum optical signal and this was also used as a loss function during the network training (see the Methods section).

After its training, we numerically tested the design of the $D^2NN$ digit classifier using 10,000 random images from MNIST test dataset (which were *not* used as part of the training or validation image sets) and achieved a classification accuracy of 91.75% (see Fig. 3C). In addition to the classification performance of the diffractive network, Fig. 3C also reports the energy distribution observed at the network output plane for the same 10,000 input test digits, the results of which clearly demonstrate that the diffractive network learned to focus the input energy of each handwritten digit into the correct (i.e., the target) detector region that it was trained for. The classification performance of this $D^2NN$ design can be further improved by e.g., increasing the number of layers, neurons and/or connections in the network.

Following these numerical results, we 3D-printed the designed 5-layer $D^2NN$ (Fig. 2A), with each layer having an area of 8 × 8 cm, followed by ten detector regions defined at the output plane of the diffractive network (Figs. 1B and 3A), and experimentally tested its classification performance using continuous wave illumination at 0.4 THz. For these experiments (Figs. 2C, D), we 3D-printed 50 handwritten digits i.e., 5 different handwritten inputs for each digit, selected among the images that numerical testing was successful, and used these objects as input to the printed diffractive network to quantify the match between numerical testing results and experiments. For each input object that is uniformly illuminated with the THz source, we imaged the output plane of the $D^2NN$ to map the intensity distribution for each detector region of interest that is assigned to a digit. Our experimental results are summarized in Fig. 3B, which demonstrates the success of the 3D-printed diffractive neural network and its learning capability: the average intensity distribution at the output plane of the network for each input digit clearly reveals that the 3D-printed $D^2NN$ was able to focus the input energy of the beam and achieve a maximum signal at the corresponding detector region that was assigned for that digit. Due to 3D-printing errors and possible alignment issues, there are also small discrepancies between our numerical simulations and experimental results for the exact distribution of the output energy for different digits. For example, as illustrated in our numerical test results (Fig. 3C), the designed $D^2NN$ was able to focus on average ~38-53% of the total output plane energy onto the correct detector region, for 10,000 different input test digits. In our experiments corresponding to 50 different 3D-printed handwritten digits, the same percentage dropped to ~34-40%, with one exception, the digit "1", focusing on average ~60% of the total output plane energy onto its correct detector region. This relatively small reduction in the performance of the experimental network is especially more pronounced for the digit "0" since it is challenging to 3D-print the large void region at the center of the digit; similar printing challenges were also observed for other digits that have void regions, e.g., "6", "8", "9". Despite 3D-printing errors, possible alignment issues and other experimental error sources in our set-up, the match between the experimental and numerical testing of our 5-layer $D^2NN$ design was found to be 88% (Fig. 3B).

**D²NN trained as an imaging lens – a physical auto-encoder.** Next, we tested the performance of a phase-only D²NN, composed of five 3D-printed transmission layers (see Fig. 2B), which was trained to create a unit-magnification image of the input optical field amplitude at its output plane (~9 × 9 cm). The training process used ~1,500 images that are randomly selected from ImageNet database (*24*) (see the Methods section for further details). As illustrated in Fig. S1(A, C), the trained network initially connects every single amplitude point at the input plane to various neurons and features of the following layers, which then focus the light back to a point at the output (i.e., image) plane, which is, as expected, quite different than the case of free-space diffraction, illustrated in Fig. S1(B, D).

After its training and blind testing, numerically proving the imaging capability of the network as shown in Figs.S1-2, next we 3D-printed this designed diffractive network. Using the same experimental set-up shown in Fig. 2(C, D), we imaged the output plane of the 3D-printed D²NN for various input objects that were uniformly illuminated by continuous wave radiation at 0.4 THz. Fig. 4 summarizes our experimental results achieved with this 3D-printed D²NN, which successfully projected unit-magnification images of the input patterns at the output plane of the network, learning the function of an imaging lens, or a physical auto-encoder. In the same figure, we also present, for comparison, the experimental results achieved over the same sample-output plane distance (29.5 mm), *without* the 3D-printed network.

To evaluate the point spread function of this D²NN, we imaged pinholes with different diameters (1 mm, 2 mm and 3 mm), which resulted in output images, each with a FWHM of 1.5 mm, 1.4 mm and 2.5 mm, respectively (Fig. 4B). Our experimental results also revealed that the printed network can resolve a line-width of 1.8 mm at 0.4 THz (corresponding to a wavelength of 0.75 mm in air), which is slightly worse in spatial resolution compared to the numerical testing phase of our D²NN design, where the network could resolve a line-width of ~1.2 mm (see Fig. S2C). This experimentally observed degradation in the performance of the diffractive network can be due to 3D-printing errors and potential misalignments between the printed layers. Furthermore, there are some absorption related losses in the 3D-printed structure that introduce residual amplitude modulation, which partially deviates from our phase-only network design, and can be an additional source of error; this also partly explains the experimental performance degradation of the handwritten digit classification network reported earlier. This point will be further discussed in the next section.

Note also that, based on the large area of the 3D-printed network layers (9 × 9 cm) and the short axial distance between the input (output) plane and the first (last) layer of the network, i.e., 4 mm (7 mm), one can infer that the theoretical numerical aperture of this system approaches 1 in air (see Fig. 2B). During the training phase, however, our diffractive network learned to utilize only part of this spatial frequency bandwidth, which should be due to the relatively large-scale of the image features that we used in the training image set (randomly selected from ImageNet database). If a higher resolution imaging system is desired, images that contain much finer spatial features can be utilized as part of the training phase to design a $D^2NN$ that can approach the theoretical diffraction-limited numerical aperture of the system. One can also change the loss function definition used in the training (see the Methods section) to teach the diffractive neural network to enhance the spatial resolution; in fact deep learning provides a powerful framework to improve image resolution by engineering the loss function used to train a neural network (*8, 13*).

## Discussion

For a $D^2NN$, after all the parameters are trained and the physical diffractive network is fabricated or 3D-printed, the computation of the network function (i.e., inference) is implemented all-optically using a light source and optical diffraction through passive components. Therefore, the energy efficiency of a $D^2NN$ depends on the reflection and/or transmission coefficients of the network layers. Such optical losses can be made negligible, especially for phase-only networks that employ e.g., transparent materials that are structured using e.g., optical lithography, creating $D^2NN$ designs operating at the visible part of the spectrum. In our experiments, we used a standard 3D-printing material (VeroBlackPlus RGD875) to provide phase modulation, and each layer of the networks shown in Fig. 2 had on average ~51% power attenuation at 0.4 THz for an average thickness of ~1 mm (see the Fig. S3). This attenuation could be further decreased by using thinner substrates or by using other materials (e.g., polyethylene, polytetrafluoroethylene) that have much lower losses in THz wavelengths. In fact, one might also use the absorption properties of the neurons of a given layer as another degree of freedom in the network design to control the connectivity of the network, which can be considered as a physical analog of the dropout rate in deep network training (*25*). In principle, a phase-only D²NN can be designed by using the correct combination of low-loss materials and appropriately selected illumination wavelengths, such that the energy efficiency of the diffractive network is only limited by the Fresnel reflections that happen at the surfaces of different layers. Such reflection related losses can also be engineered to be negligible by using anti-reflection coatings on the substrates. In our discussions so far, multiple-reflections between the layers have been neglected since such waves are much weaker compared to the directly transmitted forward-propagating waves. The match between the experimental results obtained with our 3D-printed D²NNs and their numerical simulations also supports this.

The operation principles of D²NN can be easily extended to amplitude-only or phase/amplitude-mixed transmissive or reflective designs. Whether the network layers perform phase-only or amplitude-only modulation, or a combination of both, what changes from one design to another is only the nature of the multiplicative bias terms, $t_i^l$ or $r_i^l$ for a transmissive or reflective neuron, respectively, and each neuron of a given layer will still be connected to the neurons of the former layer through a wave-interference process, $\sum_k n_k^{l-1}(x_i, y_i, z_i)$, which provides the complex-valued input to a neuron. Compared to a phase-only D²NN design, where $|t_i^l| = |r_i^l| = 1$, a choice of $|t_i^l| < 1$ or $|r_i^l| < 1$ would introduce additional optical losses, and would need to be taken into account for a given illumination power and detection SNR at the network output plane. Although not considered here since we are dealing with *passive* diffractive neural networks, one can potentially also create diffractive networks that employ a physical gain (e.g., through optical or electrical pumping, or nonlinear optical phenomena, including but not limited to plasmonics and metamaterials) to explore the domain of *amplified bias terms*, i.e., $|t_i^l| > 1$ or $|r_i^l| > 1$. At the cost of additional complexity, such amplifying layers can be useful for the diffractive neural network to better handle its photon budget and can be used after a certain number of passive layers to boost up the diffracted signal, intuitively similar to e.g., optical amplifiers used in fiber optic communication links.

Another important advantage of $D^2NN$s is that they can be easily scaled up using large-area fabrication methods (e.g., soft-lithography) and wide-field optical components and detection systems, to cost-effectively reach tens to hundreds of millions of neurons and hundreds of billions of connections in a deep neural network architecture. This provides exciting opportunities to benefit from parallel computing capabilities of optics in a scalable and power-efficient manner. For example, integration of a $D^2NN$ with lensfree on-chip imaging (*26-28*) or detection systems could provide extreme parallelism within a compact and field-portable chip-scale system with a neural network detection field-of-view of e.g., >18 cm$^2$ that can be routinely achieved using state-of-the-art CCDs merged with on-chip imaging (*29*). Such large scale all-optical diffractive neural networks operating at visible wavelengths might be transformative for various applications, including all-optical image analysis, feature detection, object classification, and might also enable new microscope or camera designs that can learn to perform unique imaging tasks using $D^2NN$s.

Some of the main sources of error in our experiments include the alignment errors, fabrication tolerances and imperfections. To mitigate these, we designed a 3D-printed holder (Figs. 2(A, B)) to self-align the multi-layer structure of a 3D-printed $D^2NN$, where each network layer and the input object were inserted into their specific slots. Based on the resolution of our 3D-printer, the misalignment error of a 3D-printed $D^2NN$ (including its holder) is estimated to be smaller than 0.1 mm compared to the ideal positions of the neurons of a given layer, and this level of error was found to have a minor effect on the network performance as illustrated in Figs. S2 and S4. In fact, a comparison of Fig. 3C (the performance of a digit classification $D^2NN$ design without any alignment errors or imperfections) and Fig. S4 reveals that the diffractive surface reconstruction errors, absorption related losses at different layers and 0.1 mm random misalignment error for each network layer, *all combined*, reduced the overall performance of the network's digit classification accuracy from 91.75% (Fig. 3C) to 89.25% (Fig. S4). This also means that some of the experimental errors that we observed in Fig. 3B can be attributed to the imperfections in 3D-printing of the handwritten digits that have a void region, e.g., "0", "6", "8" and "9".

For an inexpensive 3D-printer or fabrication method, printing/fabrication errors and imperfections, and the resulting alignment problems can be further mitigated by increasing the area of each layer and the footprint of the $D^2NN$. This way, the feature size at each layer can be increased, which will partially release the alignment requirements. The disadvantage of such an approach of printing larger diffractive networks, with an increased feature size, would be an increase in the physical size of the system and its input optical power requirements. Furthermore, to avoid bending of the network layers over larger areas, an increase in layer thickness and hence its stiffness would be needed, which can potentially also introduce additional optical losses, depending on the illumination wavelength and the material properties. In order to minimize alignment errors and improve the performance of a $D^2NN$, a *monolithic* design that combines all the layers of the network as part of a 3D fabrication method would be desirable. There are emerging 3D nano-fabrication techniques, using for e.g., laser lithography based on two-photon polymerization (*30*), that can provide an ideal solution for creating such monolithic $D^2NNs$.

Another important avenue to consider is the use of spatial light modulators (SLMs) as part of a diffractive neural network. This approach of using SLMs in $D^2NNs$ has several advantages, at the cost of an increased complexity due to deviation from an entirely passive optical network to a reconfigurable electro-optic one. First, a $D^2NN$ that employs one or more SLMs can be used to learn and implement various tasks because of its reconfigurable architecture. Second, this reconfigurability of the physical network can be used to mitigate alignment errors or other imperfections in the optical system of the network. Furthermore, as the optical network statistically fails, e.g., a misclassification or an error in its output is detected, it can mend itself through a transfer learning based re-training with appropriate penalties attached to some of the discovered errors of the network as it is being used. For building a $D^2NN$ that contains SLMs, both reflection and transmission based modulator devices can be used to create an optical network that is either entirely composed of SLMs or a hybrid one, i.e., employing some SLMs in combination with fabricated (i.e., passive) layers.

Finally, the use of the mathematical basis of D$^2$NNs in computer-based neural networks is an avenue to further explore. As discussed in our Results section (D$^2$NN Architecture), there are various differences between the D$^2$NN framework and the standard deep neural networks used in computer science. A question that remains to be addressed is whether or not a physics-inspired neural network design, *implemented entirely in a computer*, can provide additional insights and capabilities for machine learning field. Based on the laws of wave propagation and optical diffraction, can a virtual neural network improve its performance by communicating between different layers of a network with coherent or partially-coherent waves, where each neuron modules the phase and/or amplitude of this secondary wave, together with the wave interference created by the former layers. Because a 3D-printed D$^2$NN *strictly* follows the laws of physics, we were bound with a certain form of wave function (Eq. 1). In a virtual D$^2$NN that is implemented in a computer, this connection to physics can also get looser: one can select different forms/modes of wave functions, $w_i^l$, and train the deep network in a similar fashion through virtual wave propagation. In a sense, the laws of optical diffraction can be altered in a computer environment to potentially provide advantages to the network performance. One can intuitively argue that a learned feature in a standard convolutional neural network is in fact an artificial wave of its own kind, providing a convolution operation at the following layer, similar to physical wave propagation, although several mathematical differences exist. Regardless, it remains to be rigorously explored, whether or not, a D$^2$NN architecture, with its own wave-based communication mechanism between the layers of a deep network, can provide advantages over traditional implementations of deep networks in computers.

## Acknowledgements

We thank Yibo Zhang, Hongda Wang, Zoltán S. Göröcs, Deniz Mengu and Yichen Wu of UCLA for insightful discussions. **Funding:** The Ozcan Research Group at UCLA acknowledges the support of the National Science Foundation (NSF) and the Howard Hughes Medical Institute (HHMI).

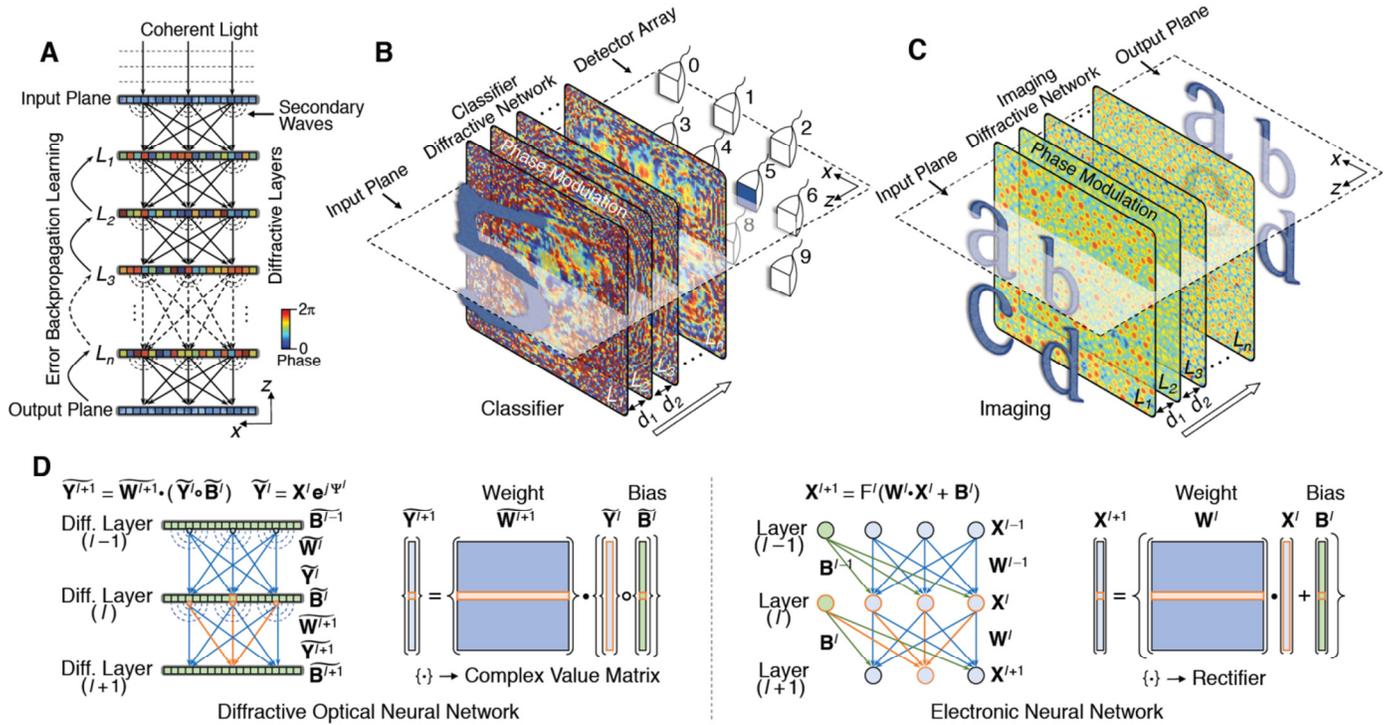

**Fig. 1: Diffractive Deep Neural Networks (D²NN).** (**A**) D²NN comprises multiple transmissive (or reflective) layers, where each point on a given layer acts as a neuron, with a complex-valued transmission (or reflection) coefficient. These transmission/reflection coefficients of each layer can be trained using deep learning to learn and perform a function between the input and output planes of the network. After this learning phase, the D²NN design is fixed, and once it is fabricated or 3D-printed, it performs the learned function at the speed of light. We trained and experimentally implemented two different D²NNs: (**B**) a handwritten digit classifier and (**C**) an imaging lens. (**D**) A comparison between D²NN and a conventional neural network is presented. Based on coherent waves, D²NN operates on complex-valued inputs, with multiplicative bias terms. Weights in a D²NN are based on free-space diffraction and determine the coherent interference of the secondary waves that are phase and/or amplitude modulated by the previous layers. "**o**" refers to a Hadamard product operation.

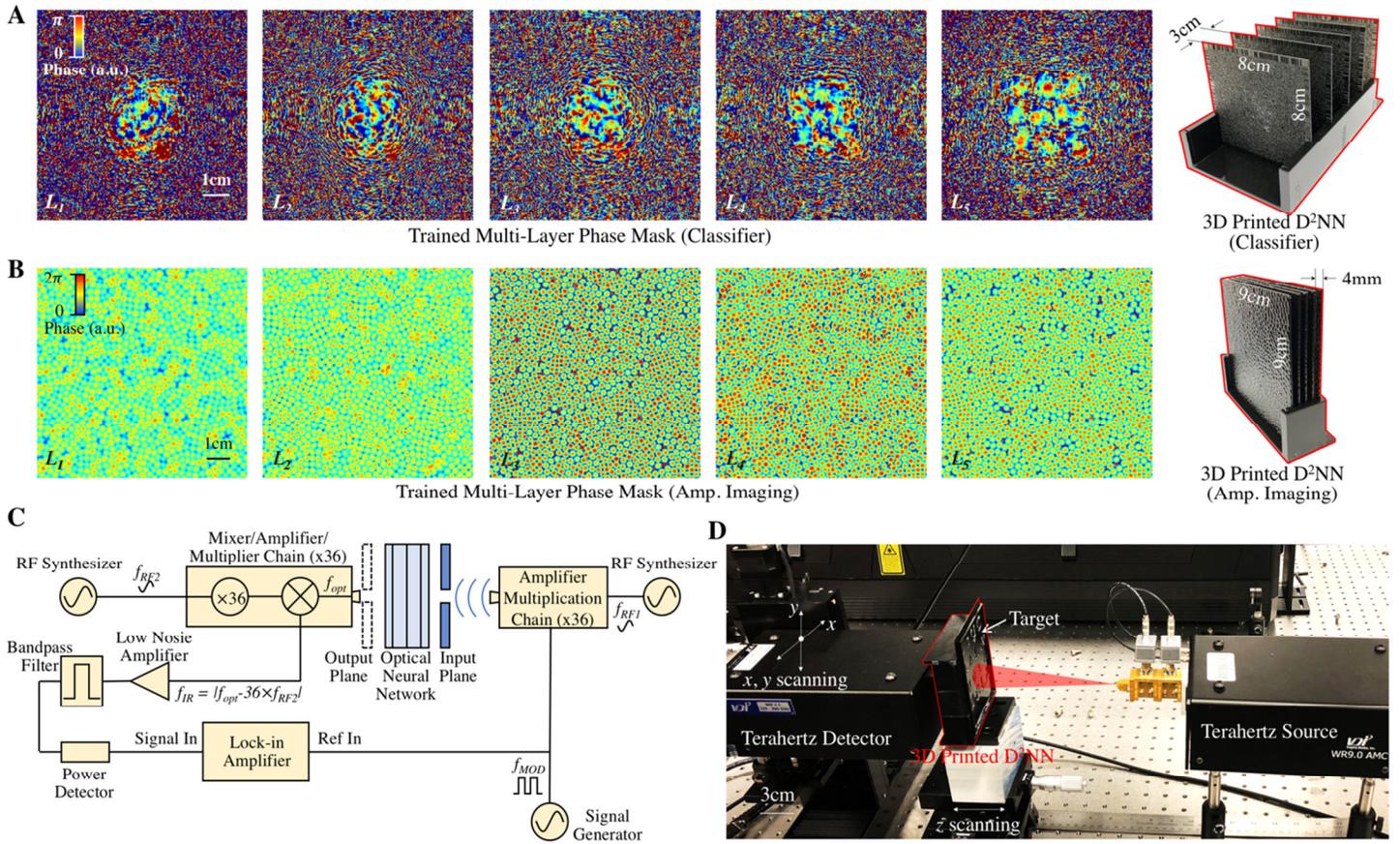

**Fig. 2: Experimental testing of 3D-printed D²NNs. (A, B)** After the training phase, the final designs of five different layers ($L_1, L_2,\ldots, L_5$) of the handwritten digit classifier and the imaging lens D²NNs are shown. To the right of each panel of the network layers, a picture of the corresponding 3D-printed D²NN is shown. A schematic (**C**) and a picture (**D**) of the experimental THz setup are shown. An amplifier/multiplier chain was used to generate continuous wave radiation at 0.4 THz and a mixer/amplifier/multiplier chain was used for the detection at the output plane of the network.

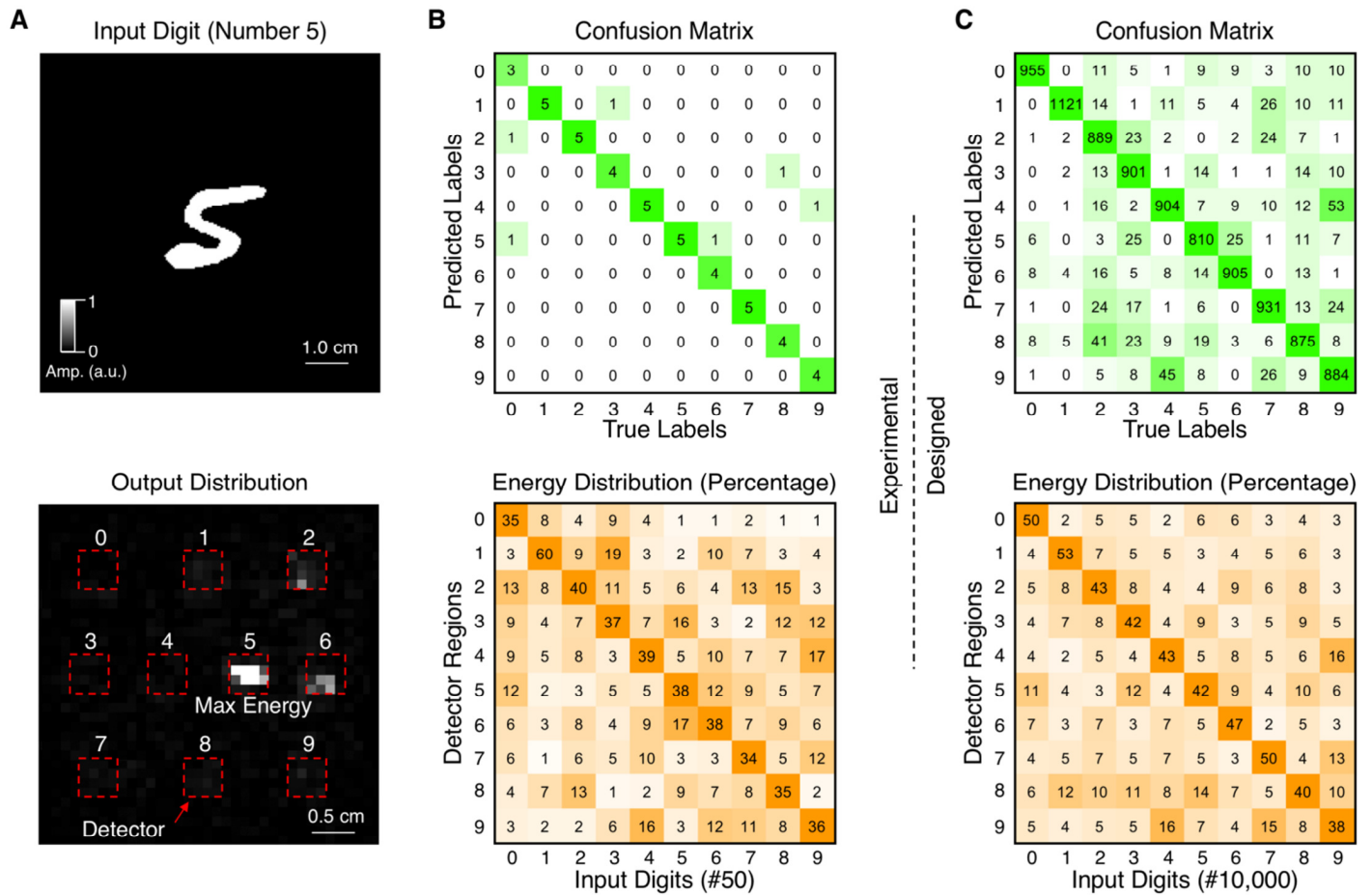

**Fig. 3: Handwritten digit classifier D²NN.** (**A**) A 3D-printed D²NN successfully classifies handwritten input digits (0, 1, …,9) based on 10 different detector regions at the output plane of the network, each corresponding to one digit. As an example, the output image of the 3D-printed D²NN for a handwritten input of "5" is demonstrated, where the red dotted squares represent the trained detector regions for each digit. Other examples of our experimental results are also shown in Fig. S5. (**B**) shows the confusion matrix and the energy distribution percentage for our experimental results, using 50 different handwritten digits that were 3D-printed (i.e., 5 for each digit), selected among the images that numerical testing was successful. (**C**) is the same as (**B**), except it summarizes our numerical testing results for 10,000 different handwritten digits (approximately 1,000 for each digit), achieving a classification accuracy of 91.75%.

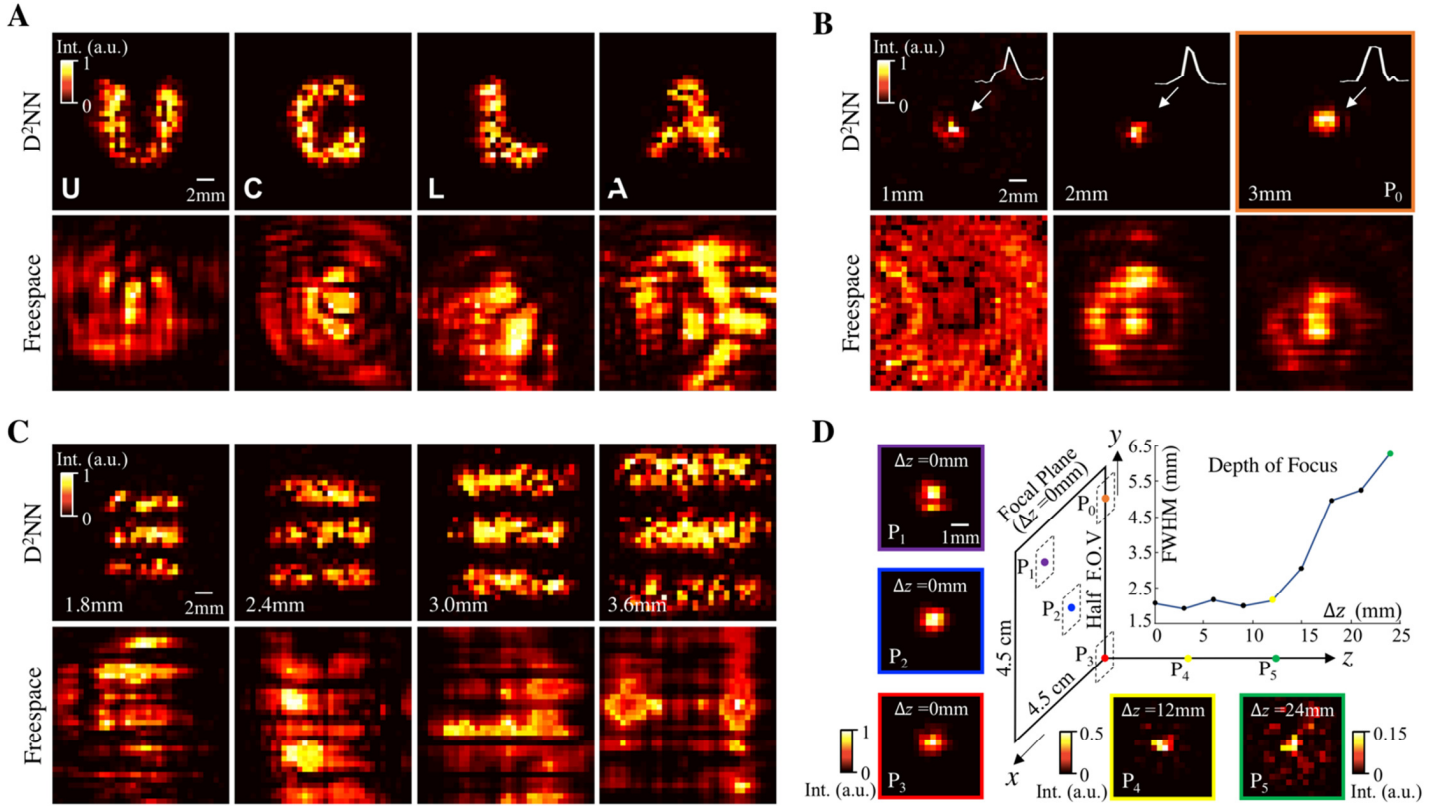

**Fig. 4: Imaging lens D²NN.** (**A**) Output images of the 3D-printed lens D²NN are shown for different input objects: 'U', 'C', 'L' and 'A'. To be able to 3D-print letter 'A', the letter was slightly modified as shown in the bottom-left corner of the corresponding image panel. For comparison, free-space diffraction results corresponding to the same objects, without the 3D-printed D²NN, are also shown. (**B**) Same as in (**A**), except the input objects were pinholes with diameters of 1 mm, 2 mm and 3 mm. (**C**) D²NN can resolve a line-width of 1.8 mm at its output plane. (**D**) Using a 3-mm pinhole that is scanned in front of the 3D-printed network, we evaluated the tolerance of the physical D²NN as a function of the axial distance. For four different locations on the input plane of the network, i.e., $P_1$-$P_3$, in (D) and $P_0$ in (B), we obtained very similar output images for the same 3-mm pinhole. The 3D-printed network was found to be robust to axis defocusing up to ~12 mm from the input plane.